\documentclass[journal]{IEEEtran}

\usepackage[table,xcdraw]{xcolor}

\usepackage[utf8]{inputenc}
\usepackage[colorinlistoftodos,prependcaption,textsize=tiny]{todonotes}

\usepackage{url}
\usepackage{amsmath}
\usepackage{cite}
\usepackage{booktabs}

\begin{document}

%
\title{Injury risk prediction for traffic accidents \\ in Porto Alegre/RS, Brazil (draft)}

\author{Christian S. Perone, \IEEEmembership{Pontifícia Universidade Católica do Rio Grande do Sul (PUC-RS).}}

\maketitle

\begin{abstract}
This study describes the experimental application of Machine Learning techniques to build prediction models that
can assess the injury risk associated with traffic accidents. This work uses an freely available
data set of traffic accident records that took place in the city of Porto Alegre/RS (Brazil) during the
year of 2013. This study also provides an analysis of the most important attributes of a traffic
accident that could produce an outcome of injury to the people involved in the accident.
\end{abstract}


\IEEEpeerreviewmaketitle

\section{Introduction}
Statistics provided by EPTC \cite{EPTC:Statistics} -- the traffic managing agency in Porto Alegre/RS (Brazil) --
shows that in 2013, approximately 22.447 traffic accidents took place in the city of Porto Alegre,
an average of 1.870 traffic accidents per month. According to Saunier et al. \cite{saunier}, the social cost of road
collisions is the largest side effect
of road transportation . The costs of fatalities, injuries and property damage, as well as medical
care and traffic delays accounts for a significant impact on the finance of the people involved, cities and the
government. According to Brazilian National Traffic Department (DENATRAN) \cite{DENATAN:CustosAcidentes}, the
average cost for an accident (in federal highways) without victims is R\$ 16.840,00, for accidents with victims
this cost increases to R\$ 86.032,00 and for accidents with fatalities, the cost is R\$ 418.341,00. 

Descriptive analysis of the situation in road safety and road accidents are important, but understanding 
the factors related with dangerous situations and patterns in data is of even greater importance \cite{Beshah}.
Being able to predict when an traffic accident will result in an injury, can help traffic agencies to
provide faster medical care.
Another example of the benefits of understanding the factors behind the
injury risk is to guide traffic agencies to improve the road safety by means of infrastructure design (which
includes road signs and speed control devices) or even through the pedestrian/driver behavior improvements that could be obtained
with targeted marketing campaigns.
Data-driven decisions can also help the traffic agencies to reduce
the costs involved in a traffic accident.

This paper describes the efforts and experimental results obtained through the application of Machine Learning
techniques in order to provide a better understanding of the data that is being collected today by the traffic
agency. The main purpose of this work, is to get an overall understanding of the accident data as well as
build a predictor for the injury risk related to the traffic accidents in the city of Porto Alegre/RS (Brazil). 

This study is organized as follows: in Section \ref{related_work}, an overview of the related work is provided, followed
by the Section \ref{proposed_approach}, where an overview of our proposed approach
to the problem is described. In Section \ref{experimental_methodology}, the experimental methodology is provided,
including which algorithms were used an how they were used, while in Section \ref{experimental_analysis} the
results of the experimental analysis are detailed followed by the Section \ref{conclusion}, that describes the
conclusions and remarks related to future work.

\section{Related Work}
\label{related_work}
An attempt has been made to search for existing accident analysis practice in the city of Porto Alegre, however
no published works were found related to the application of Machine Learning techniques by the traffic managing 
agency (EPTC) in Porto Alegre. Only limited descriptive analysis were published
on the site of the traffic agency \cite{EPTC:Statistics}, but no deeper analysis of the factors or
injury risk assessment were published by the traffic managing agency.
The lack of standardization for the data collection process and for the data itself between different
traffic managing agencies worldwide makes the experimental results comparison very limited.

Beshah et al. \cite{Beshah} explored a rich data set, comprising of 14.254 accident cases described with
48 attributes containing information related to road users (drivers, pedestrians and passengers), vehicles
and road. In their study, two predictive modeling methods were used: CART and Random Forests. The
experimental results done using CART analysis to assess the injury risk, scored with respect to the
area under the ROC curve (AUC) a result of 0.8827. While running Random Forests, the authors also found
that the age of the victim, victim occupation, among others, were the attributes with the most predictive power.

Saunier et al. \cite{saunier} investigated the collision factors and processes (i.e. the chain of events
that lead to collisions) through the collection and analysis of microscopic data (road users trajectories).
Saunier et al. \cite{saunier} avoided the use of algorithms with a ``black box'' nature like ANNs (Artificial
Neural Networks) or SVMs (Support Vector Machines) and used C4.5 (Decision Trees) instead and clustering
analysis to investigate the collision factors. In their work, they found an strong relationship between the
evasive actions and the interaction outcome: in most collisions (62 out of 82), no evasive action was
attempted \cite{saunier}.

\section{Proposed Approach}
\label{proposed_approach}
The work described in this paper aims to evaluate different Machine Learning techniques in order to build
a predictive model for injury risk assessment of traffic accident events based on data that was collected
by a traffic managing agency. Since the injury risk has a dichotomous nature in relation to the dependent variable,
this facilitates the use of binary classifiers used in this study like Logistic Regression or Support Vector Machines.
This paper also evaluates the association between the traffic accident injury outcome and the possible
contributory factors, an effort to understand which are the most important factors in an accident with
an injury outcome.

\section{Experimental Methodology}
\label{experimental_methodology}
This section describes the data set used, as well as the tools and algorithms used to perform the analysis.

\subsection{Traffic Accident Data Set}
The data set used in this study was obtained through Datapoa \cite{Datapoa}, an initiative from the city hall
of Porto Alegre to provide open data access to many data sets related with the city itself. The traffic accident
data set available at the Datapoa is licensed under the Open Database License (ODbL) \cite{ODBL}, which is an Attribution
and Share-Alike license for databases.

Although the time span of the available traffic accident data sets ranges from the year 2000 up to 2013, only the data from
the most recent data set was used (the data set related to the accidents that happened in the year of 2013). The data set
is comprised of 20.798 accident records described using 44 attributes. Some attributes of the data set are irrelevant
for the purpose of this study and many attributes also presented duplicated data or invalid records, thus a step of data
cleansing was required before using the data set. The data set also lacks detailed information about vehicles (i.e. age, movement),
drivers (i.e. age of the driver, driver license level, driving experience, sex, etc.) and victims (i.e. age).

\subsection{Tools}
To plot heat maps with the geospatial distribution of the accidents, this study used the framework
Django GIS Brasil \cite{DjangoGISBrasil}, an open source project from the same author of this study, 
that aggregates geospatial information related with the Brazilian territory. To provide data analysis,
the author used Pandas \cite{Pandas}, an open source library providing high-performance, data structures
and data analysis tools for the Python programming language. This study also used
scikit-learn \cite{scikit-learn} -- an open source Machine Learning framework for the Python language --
to perform data pre-processing and to build the predictive models.

\subsection{Machine Learning Techniques}
This study employed the following algorithms as classifiers for the injury risk assessment: Logistic Regression,
Support Vector Machines, Naive Bayes, K-nearest neighbors and Random Forests. The details about the use,
parametrization and model evaluation techniques used to assess the predictive models are described
in the next sections.

\subsection{Logistic Regression}
The Logistic Regression used in this study is the Logistic Regression present in the scikit-learn framework \cite{scikit-learn},
which in turn uses the LIBLINEAR \cite{Liblinear} implementation of the Logistic Regression. The LIBLINEAR implementation
solve the following optimization problem:
\begin{equation}
\min_{\substack{w}} \frac{1}{2} w^Tw + C \sum_{i=1}^{l} \xi (w; x_i, y_i)
\end{equation}
Given a set of instance-label pairs $(x_i, y_i), i = 1, \ldots, l$ where $C$ is the penalty parameter and $\xi (w; x_i, y_i)$
is the loss function, which for Logistic Regression is:
\begin{equation}
log(1 + e^{-y_i w^T x_i})
\end{equation}
In this study we used L2 regularized Logistic Regression with the penalty $C$ equal to 1.0.

\subsection{Support Vector Machines}
The Support Vector Machine (SVM) from scikit-learn \cite{scikit-learn} used in this work is based on the LIBSVM \cite{Libsvm} implementation,
which is a $C$-Support Vector Classification. For more details about the algorithm implementation, please refer to the elucidative LIBSVM \cite{Libsvm}
original paper. The SVM algorithm was parametrized with a linear kernel and with 9.0 as the error term, both parameters were chosen using 
hyperparameter optimization through a non-exhaustive grid search between different kernel types (RBF, Polynomial and Linear) with different
error term and gamma values. It is also important to note that Support Vector Machine algorithms are not scale invariant, so the the author
applied a scaling function over the attributes before feeding attributes into the algorithm.

\subsection{Naive Bayes}
The Naive Bayes algorithm used in this study is also from scikit-learn \cite{scikit-learn}. The different
Naive Bayes classifiers implement in scikit-learn differ mainly by the assumptions they make regarding the
distribution of $P(x_i \mid y)$ \cite{scikit-learn}. The author decided to use the Gaussian Naive Bayes, 
where the likelihood of the features is assumed to be Gaussian:
\begin{equation}
P(x_i \mid y) = \frac{1}{\sqrt{2\pi\sigma^2_y}} \exp\left(-\frac{(x_i - \mu_y)^2}{2\sigma^2_y}\right)
\end{equation}
And where the parameters $\sigma_y$ and $\mu_y$ are estimated using maximum likelihood \cite{scikit-learn}.

\subsection{K-nearest neighbors}
The K-nearest neighbors (kNN) algorithm used in this study is also from scikit-learn \cite{scikit-learn}, which 
provides both unsupervised and supervised neighbors-based learning methods. Despite the simplicity of the
algorithm, kNN has been successful in a large number of classification and regression problems. The number
of neighbors used in this study (for the $k$ value) is 8. This value was also found using a non-exhaustive
hyperparameter optimization through the grid search technique. Attribute scaling was also performed before
using kNN, to ensure that the distance measure accords equal weight to each variable.

\subsection{Random Forests}
The author of the paper also used Random Forests from scikit-learn \cite{scikit-learn} as a binary classifier
and also to evaluate the feature importance in order to understand which are the most important factors while
predicting the injury risk. Random forests are a combination of tree predictors, where each tree
depends on the values of a random vector sampled independently and with the same distribution for
all trees in the forest \cite{RandomForest}.
When compared with the original publication from Brainman \cite{RandomForest}, the scikit-learn implementation
combines classifiers by averaging their probabilistic prediction, instead of letting each classifier vote
for a single class \cite{sklearn-randomforest}. Random Forests were also used to assess injury risk and the importance
of factors in the aforementioned study done by Beshah \cite{Beshah}.

The number of estimators (trees in the forest) used was 200, this number was chosen using non-exhaustive
hyperparameter optimization through grid search.

\subsection{Model Evaluation}
In order to evaluate the predictive models trained in this study, the author used a cross-validation with a
train data set with 60\% of the instances from the original data set and with a test data set with 40\%
of the original data set. Both the training and the testing data set were random sampled from the original
data set.

To evaluate the predictive models, the author used the Area Under the Curve (AUC) -- \textit{computed using the
trapezoidal rule} -- of the Receiver Operating Characteristic (ROC), which is a graphical plot that shows the performance of a
binary classifier varying the discrimination threshold. The curve in this study was plotted
using the true positive rate against the false positive rate at various threshold settings (one for
each different predictive outcome from each model).

Also, complementary to the ROC and AUC, the author of this study calculated tables presenting the Precision, Recall
and F1-Score for each class from each predictive model used.

Since some algorithms used in this study didn't have a natural probability estimate outcome like Logistic Regression has
for each class, the author hence used different estimating techniques
in order to be able to compare the ROC and AUC between different classifiers:

\subsubsection{Support Vector Machines} In this case, the probability estimates were calculated by LIBSVM using Platt scaling \cite{Platt}.
\subsubsection{K-nearest neighbors} For kNN, the predicted probability for each class is the ratio of neighbors voting for each label, i.e.
if $k=5$ and 4 neighbors predicted class 1 and only one neighbor predicted class 0, then the probabilities for that example is 0.2 and 0.8.
\subsubsection{Random Forest} The probabilities of a forest are the mean probabilities of the trees in the ensemble and the probabilities
returned by a single tree are the normalized class histograms of the leaf that a sample lands in.

\subsection{Data Cleansing}
Some accident records present in the data set had an invalid date/time format and since the amount of invalid records was
very low (less than 5) when compared with the amount of valid records (greater than 20.790), these records were just
removed from the data set without causing any significant loss in the experimental analysis.

\section{Experimental Analysis}
\label{experimental_analysis}
This section provides a brief descriptive analysis of the data set used as well as the experimental results
using different predictive models together with their model evaluations.

\subsection{Data set analysis}
The data set (after applying data cleansing), is comprised by 20.798 records of traffic accident events that took place in
the city of Porto Alegre/RS (Brazil). The attributes of the data set can be categorized in TODO different types:

\subsubsection{Geospatial Attributes}
These attributes, listed in Table \ref{table:geospatial_attributes} represents where the accident happened in space.
They weren't used in this study and were left for a further study.

\begin{center}
\begin{table}[h]
\renewcommand{\arraystretch}{1.3}
\caption{Geospatial Attributes}
\label{table:geospatial_attributes}
\centering
\begin{tabular}{ll}
\hline
Attribute Name         & Description                               \\ \hline
LOG1 and LOG2          & Street names.                              \\
PREDIAL1               & Street numbers.                            \\
REGION                 & Region of the city.                        \\
LATITUDE and LONGITUDE & The geographical coordinates.              \\
LOCAL\_VIA and REGION  & Concatenation of LOG1, LOG2 and PREDIAL1.
\end{tabular}
\end{table}
\end{center}

\subsubsection{Irrelevant Attributes}
These attributes are irrelevant to the analysis of factors or injury risk assessment.
They are presented in Table \ref{table:irrelevant_attributes}.

\begin{center}
\begin{table}[h]
\renewcommand{\arraystretch}{1.3}
\caption{Irrelevant Attributes}
\label{table:irrelevant_attributes}
\centering
\begin{tabular}{ll}
\hline
Attribute Name         & Description                               \\ \hline
ID                     & The unique ID of the accident.             \\
BOLETIM                & The ID of the traffic agency record.
\end{tabular}
\end{table}
\end{center}

\subsubsection{Attributes with data leakage}
Since the main goal of this study is to predict the risk of injury/non-injury, an extra care was taken to discover
attributes that could leak to the target class. The result of this evaluation is present in the Table \ref{table:leak_attributes}.
The attribute ``FONTE'' leaks information about the injury target class because the police is usually involved only when
there was someone injuried. The attribute ``UPS'' also leaks information about the target class because it assumes 3 different
values: 1 (accident only with property damage), 5 (accident with someone injuried) and 13 (accident with deaths), so when the
UPS is 5 or 13 it will perfect predict the injuried/non-injuried target classes.

\begin{center}
\begin{table}[h]
\renewcommand{\arraystretch}{1.3}
\caption{Attributes with data leakage}
\label{table:leak_attributes}
\centering
\begin{tabular}{ll}
\hline
Attribute Name         & Description                               \\ \hline
FONTE                  & Whether the accident was registered by the traffic managing \\
                       & agency or by the police. \\
UPS                    & A severity measurement.
\end{tabular}
\end{table}
\end{center}

\subsubsection{Relevant attributes}
These are attributes that were used to train all the predictive models presented in this study. They are shown
in the Table \ref{table:attributes_relevant}. Except the counting attributes, all attributes were preprocessed
using one-hot encoding scheme (aka. one-of-K scheme).

\begin{center}
\begin{table}[h]
\renewcommand{\arraystretch}{1.3}
\caption{Relevant Attributes}
\label{table:attributes_relevant}
\centering
\begin{tabular}{ll}
\hline
Attribute Name         & Description                               \\ \hline
LOCAL                  & Whether the accident happened on a street or in crossing streets. \\
TIPO\_ACID              & The type of the accident (collision, fire, etc...). \\
DIA\_SEM								 & The day of the week. \\
CONSORCIO						   & If a bus were involved, the name of the company. \\
AUTO								   & The count of cars involved. \\
TAXI								   & The count of cabs involved. \\
LOTACAO								 & The count of small bus involved. \\
ONIBUS\_URB						 & The count of urban bus involved. \\
ONIBUS\_MET						 & The count of bus (others) involved. \\
CAMINHAO							 & The count of trucks involved. \\
MOTO							     & The count of motorcycles involved. \\
CARROCA						   	 & The count of carts involved. \\
BICICLETA					 		 & The count of bikes involved. \\
OUTRO							     & The count of vehicles (others) involved. \\
TEMPO						     	 & How was the weather (raining, clear, etc.). \\
NOITE\_DIA							 & If it was night or day. \\
MES							       & The month of the accident. \\
FX\_HORA							   & The hour that accident happened. \\
CORREDOR							 & Whether the accident happened in the bus lane road or not.
\end{tabular}
\end{table}
\end{center}

\subsubsection{Target attribute}
Since the aim of this work is to predict if the outcome of an traffic accident was an injury/non-injury,
the author merged (summed) the features shown in the Table \ref{table:target_attributes} and then
created a new attribute with this value that was later converted to 0 (non-injury) if sum was less or equal than
zero, or 1 (injury) if the sum was greater or equal to 1.

\begin{center}
\begin{table}[h]
\renewcommand{\arraystretch}{1.3}
\caption{Attributes merged to create the target attribute}
\label{table:target_attributes}
\centering
\begin{tabular}{ll}
\hline
Attribute Name         & Description                               \\ \hline
FERIDOS                & The count of injured people involved in the accident. \\
FERIDOS\_GR             & The count of serious injured people involved in the accident. \\
MORTES                 & The count of deaths (local deaths) in the accident. \\
MORTES\_POST            & The count of deaths (posterior deaths) happened after the accident. \\
FATAIS                 & The sum of MORTES and MORTES\_POST attributes.
\end{tabular}
\end{table}
\end{center}

It is also important to note that the data set is imbalanced and it has a ratio of records of at least 2:1
between the target classes (injury/non-injury), totaling 14.247 non-injury instances and 6.551 injury 
records.

\begin{figure}[!t]
\centering
\includegraphics[width=3.5in]{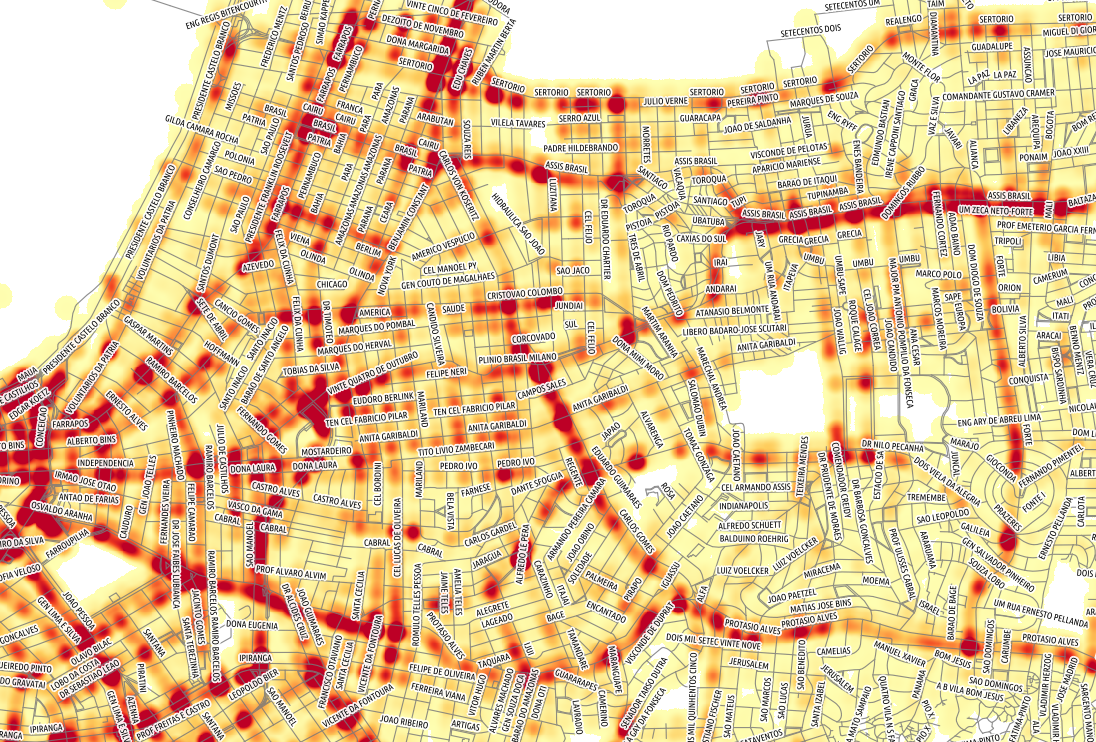}
\caption{Heat map of the traffic accidents.}
\label{figure:heatmap}
\end{figure}

The geospatial information related to the accident events weren't used in this study, but the heat map shown in Figure
\ref{figure:heatmap} shows an important pattern that clearly confirms that the accidents density increases
on crossing streets. This information is represented not only in latitude/longitude attributes but also in
the ``LOCAL'' attribute used to train the predictive models in this study.

\subsection{Model evaluation}
\begin{figure}[!t]
\centering
\includegraphics[width=3.6in]{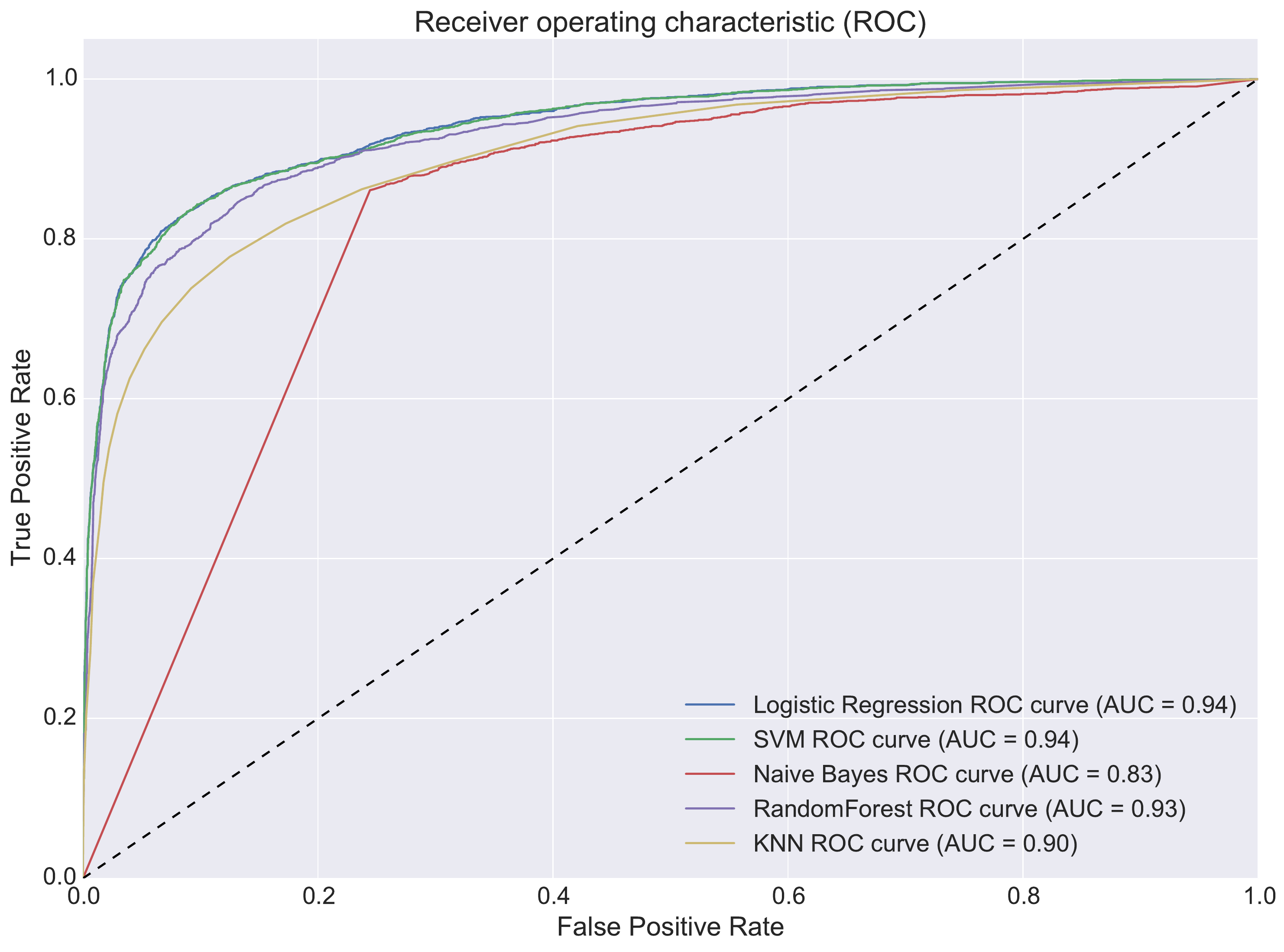}
\caption{The Receiver Operating Characteristic (ROC) and Area Under Curve (AUC) of each model for the positive class (injury).}
\label{figure:roc}
\end{figure}

The model evaluation results (Precision, Recall, F1-score) using test data set for Logistic Regression and Support
Vector Machine (SVM) are shown in the Table \ref{table:svm_log}, the results for Naive Bayes
and K-nearest neighbors are shown in the Table \ref{table:bayes_knn} and the results for the
Random Forest model evaluation is shown in the Table \ref{table:random_forest}. The ROC curves and the AUC
value for each model for the positive class (injury) prediction with varying discriminative threshold is presented
in the Figure \ref{figure:roc}.

As we can see, Logistic Regression and Support Vector Machines models provided the best scores
in AUC and average Precision/Recall/F1-score, they also were very similar regarding the performance. Random Forest also performed
well with an AUC of 0.93 when compared with AUC of SVM and Logistic Regression that scored an AUC of 0.94 each.
K-nearest neighbor performed below the scores of SVM, Logistic Regression and Random Forest with an AUC of 0.90, but
it still performed better than the worst model which is Naive Bayes with an AUC of 0.83.

\begin{center}
\begin{table}[h]
\renewcommand{\arraystretch}{1.2}
\caption{Support Vector Machine and Logistic Regression evaluation}
\label{table:svm_log}
\centering
\begin{tabular}{llll|lll}
\hline
\multicolumn{4}{c}{Support Vector Machine} & \multicolumn{3}{c}{Logistic Regression} \\ \hline
           & Precision & Recall & F1-score & Precision     & Recall     & F1-Score    \\ \hline
Non-injury & 0.90      & 0.96   & 0.93     & 0.90          & 0.96       & 0.93        \\
Injury     & 0.89      & 0.76   & 0.82     & 0.89          & 0.76       & 0.82        \\ \hline
Average    & 0.90      & 0.90   & 0.89     & 0.89          & 0.90       & 0.89     
\end{tabular}
\end{table}
\end{center}

\begin{center}
\begin{table}[h]
\renewcommand{\arraystretch}{1.2}
\caption{Naive Bayes and K-nearest neighbors evaluation}
\label{table:bayes_knn}
\centering
\begin{tabular}{llll|lll}
\hline
\multicolumn{4}{c}{Naive Bayes} & \multicolumn{3}{c}{K-nearest neighbors evaluation} \\ \hline
           & Precision & Recall & F1-score & Precision     & Recall     & F1-Score    \\ \hline
Non-injury & 0.96      & 0.23   & 0.38     & 0.85          & 0.96       & 0.90        \\
Injury     & 0.37      & 0.98   & 0.54     & 0.88          & 0.63       & 0.73        \\ \hline
Average    & 0.78      & 0.47   & 0.43     & 0.86          & 0.86       & 0.85     
\end{tabular}
\end{table}
\end{center}

\begin{center}
\begin{table}[h]
\renewcommand{\arraystretch}{1.2}
\caption{Random Forest evaluation}
\label{table:random_forest}
\centering
\begin{tabular}{llll}
\hline
\multicolumn{4}{c}{Random Forest} \\ \hline
           & Precision & Recall & F1-score \\ \hline
Non-injury & 0.90      & 0.94   & 0.92     \\
Injury     & 0.85      & 0.76   & 0.80     \\ \hline
Average    & 0.88      & 0.88   & 0.88  
\end{tabular}
\end{table}
\end{center}

\subsection{Variable Importance}
According to Strobl et al. \cite{random_bias}, Random Forests have been successfully applied to various problems and within a very
short period of time, random forests have become a major data analysis tool, that performs well in comparison with many standard methods.
One of the factors that greatly contributed to the popularity of Random Forests was that it produces variable
importance measures for each predictor variable.

The experimental results to evaluate the attributes that had more importance while predicting the injury risk were obtained using
the same trained model with the evaluation presented in the Table \ref{table:random_forest}. The first 10 most important attributes that
were described by this model, with their respective importances, are presented in the Table \ref{table:attribute_importance}.

while predicting the injury risk, followed by other attributes, like cars involved and if accident was a run over, among others.
As we can see in the Table \ref{table:attribute_importance}, the motorcycle count attribute had the largest importance

\begin{center}
\begin{table}[h]
\renewcommand{\arraystretch}{1.3}
\caption{Attribute Importances}
\label{table:attribute_importance}
\centering
\begin{tabular}{llll}
Importance & Attribute Name             & Description                                 \\ \hline
\hline
0.2108     & MOTO                       & The count of motorcycles involved.             \\
0.0948     & AUTO                       & The count of cars involved.                        \\
0.0925     & TIPO\_ACID \\ 
           & ATROPELAMENTO  & If the type of the accident was a run over.                     \\
0.0391     & LOCAL \\
           & LOGRADOURO         & If the accident was on a normal street.              \\
0.0368     & LOCAL \\ 
           & CRUZAMENTO         & If the accident was on crossing streets.           \\
0.0267     & TIPO\_ACID \\
           & COLISÃO        & If the type of the accident was a collision.                        \\
0.0205     & TIPO\_ACID \\
           &  QUEDA          & If the type of the accident was a fall.                 \\
0.0203     & CAMINHAO          & The count of trucks involved.                        \\
0.0182     & TIPO\_ACID \\
           & ABALROAMENTO   & If the type of the accident \\ 
					                  & & was a collision (on the side).                  \\
0.0181    & NOITE\_DIA \\
          & DIA   & If the accident happened during night time.                          \\
\end{tabular}
\end{table}
\end{center}

\section{Conclusions and Future Work}
\label{conclusion}
As we can see, the experimental results demonstrated that prediction models for 
injury risk assessment can be created with
good precision, even with limited data sets, like the one used in this study that lacks
information about vehicle drivers, victims and vehicle movements.
These results, together with the variable importance analysis, can be used by traffic
managing agencies to understand the provided data sets with an even greater depth than the limited
descriptive analysis that is being carried today by these agencies.

This study didn't used the geospatial data, but the author believes that this information is also a
critical factor to the prediction of the injury risk associated with an traffic accident. The use of
the geospatial data was left to a future study due to the very specific nature of the geospatial
data format, which requires different preprocessing approach before being employed.

Future works can also include better hyperparameter optimization with a more intensive search
for better parameters such as kNN neighbor size, SVM error term, SVM kernel parameters, Random Forest
estimators count, among others.
This study also didn't applied feature selection techniques, but the author believes that a
future work could also improve the models performance by using feature selection methods.

\bibliography{IEEEabrv,citedatabase}

\end{document}